\documentclass[10pt,twocolumn,letterpaper]{article}

\usepackage{cvpr}
\usepackage{times}
\usepackage{epsfig}
\usepackage{graphicx}
\usepackage{amsmath}
\usepackage{amssymb}

\usepackage{subfigure}
\usepackage{dsfont}
\usepackage{color}

\usepackage{multirow}
\usepackage{tabularx}
\usepackage{arydshln}
\usepackage{mathtools}

\newcolumntype{C}[1]{>{\centering\let\newline\\\arraybackslash\hspace{0pt}}p{#1}}
\newcolumntype{L}[1]{>{\raggedright\let\newline\\\arraybackslash\hspace{3pt}}p{#1}}

\newcommand{\prob}{\text{p}}
\newcommand{\vara}{\mathbf{a}}
\newcommand{\varb}{\mathbf{b}}
\newcommand{\vard}{\mathbf{d}}

\newcommand{\varv}{v}
\newcommand{\varq}{\mathbf{q}}
\newcommand{\varr}{\mathbf{r}}
\newcommand{\vart}{\mathbf{t}}

\newcommand{\varI}{\mathbf{I}}
\newcommand{\real}{\mathbb{R}}
\newcommand{\vqa}{\text{vqa}}
\newcommand{\pre}{\text{pre}}

\newcommand{\hyeonwoo}[1]{{\color{red}{HW: #1}}}
\newcommand{\jh}[1]{{\color{blue}{#1}}}

\usepackage{url}

\DeclareMathOperator*{\argmax}{arg\!\max}

\usepackage[pagebackref=true,breaklinks=true,letterpaper=true,colorlinks,bookmarks=false]{hyperref}

\cvprfinalcopy 

\def\cvprPaperID{4024} 

\ifcvprfinal\fi
\begin{document}

\title{Transfer Learning via Unsupervised Task Discovery \\ for Visual Question Answering}

\author{
Hyeonwoo Noh$^{1,3}$\hspace{1.5cm} Taehoon Kim$^2$\thanks{This work is performed while at Devsisters.}\hspace{1.5cm} Jonghwan Mun$^{1,3}$\hspace{1.5cm} Bohyung Han$^3$\\
\hspace{-0.7cm}
\begin{tabular}{c c c}
~~$^1$Computer Vision Lab.~~ & ~~$^2$OpenAI~~ & ~~$^3$Computer Vision Lab., ECE \& ASRI~~\\
~POSTECH, Korea & ~USA & ~Seoul National University, Korea
\end{tabular}\\
\begin{tabular}{c c c}
{\tt\small$^1$\{shgusdngogo,jonghwan.mun\}@postech.ac.kr} & {\tt\small $^2$carpedm20@gmail.com} & {\tt\small$^3$bhhan@snu.ac.kr} 
\end{tabular}
}

	\maketitle

\begin{abstract} 
We study how to leverage off-the-shelf visual and linguistic data to cope with out-of-vocabulary answers in visual question answering task.
Existing large-scale visual datasets with annotations such as image class labels, bounding boxes and region descriptions are good sources for learning rich and diverse visual concepts.
However, it is not straightforward how the visual concepts can be captured and transferred to visual question answering models due to missing link between question dependent answering models and visual data without question.
We tackle this problem in two steps: 1) learning a task conditional visual classifier, which is capable of solving diverse question-specific visual recognition tasks, based on unsupervised task discovery and 2) transferring the task conditional visual classifier to visual question answering models.
Specifically, we employ linguistic knowledge sources such as structured lexical database (\textit{e.g.} WordNet) and visual descriptions for unsupervised task discovery, and transfer a learned task conditional visual classifier as an answering unit in a visual question answering model.
We empirically show that the proposed algorithm generalizes to out-of-vocabulary answers successfully using the knowledge transferred from the visual dataset.
\end{abstract}


\section{Introduction}

Human sees and understands a visual scene based on diverse visual concepts.
For example, from a single image of a chair, human effortlessly recognizes diverse visual concepts such as its color, material, style, usage, and so on.
Such diverse visual concepts may be associated with different questions in natural languages defining a recognition task for each of the visual concepts ({\it e.g.}, what color is the chair?).
Recently visual question answering (VQA)~\cite{antol2015vqa} is proposed as an effort to learn deep neural network models with capability to perform diverse visual recognition tasks defined adaptively by questions.

Approaches to VQA rely on a large-scale dataset of image, question and answer triples, and train a classifier taking an image and a question as inputs and producing an answer.
Despite recent remarkable progress~\cite{anderson2017updown, fukui2016multimodal, yang2016stacked}, this direction has a critical limitation that image, question and answer triples in datasets are the only source for learning visual concepts.
Such drawback may result in lack of scalability because the triplets may be collected artificially by human annotators with limited quality control and have weak diversity in visual concepts.
In fact, VQA datasets~\cite{agrawal2017don, goyal2017making} suffer from inherent bias, which hinders learning true visual concepts from the datasets.

On the contrary, human answers a question based on visual concepts learned from diverse sources such as books, pictures, videos, and personal experience that are not necessarily associated with target questions.
Even for machines, there exist more natural and scalable sources for learning visual concepts: image class labels, bounding boxes and image descriptions. 
Such information is already available in large-scale~\cite{Imagenet, openimages, krishna2017visual} and can scale further with reasonable cost~\cite{papadopoulos2016we,papadopoulos2017extreme}.
This observation brings a natural question; can we learn visual concepts without question annotations and transfer them for VQA?

\begin{figure*}[t]
	\centering
	\includegraphics[width=0.98\linewidth] {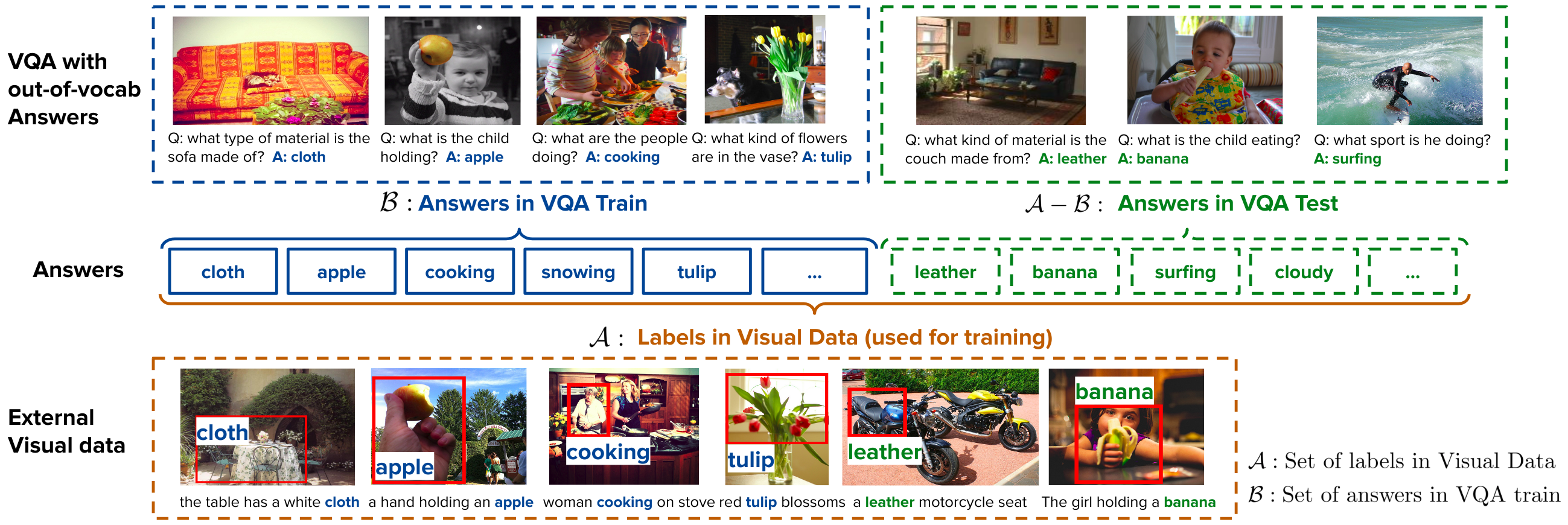}
	\vspace{0.05cm}
	\caption{
		\textbf{VQA with out-of-vocabulary answers.}
		Given a set of labels in visual dataset $\mathcal{A}$ and a set of answers in VQA training set $\mathcal{B}$, we evaluate a model on VQA test set with answers $\vara \in \mathcal{A} - \mathcal{B}$.
		External visual dataset provides a set of bounding box labels and visual descriptions for all answers in VQA training set and test set. See text for details.
	}
	\vspace{-0.3cm}
	\label{fig:zeroshot_setting}
\end{figure*}

To address this question, we introduce a VQA problem with out-of-vocabulary answers, which is illustrated in Figure~\ref{fig:zeroshot_setting}.
External visual dataset provides a set of labels $\mathcal{A}$ and only a subset of these labels $\mathcal{B} \subset \mathcal{A}$ appears in VQA training set as answers. 
The goal of this task is to handle out-of-vocabulary answers $\vara \in \mathcal{A} - \mathcal{B}$ successfully by exploiting visual concepts learned from external visual dataset.
%

This paper studies how to learn visual concepts without questions and how to transfer the learned concepts to VQA models.
To learn transferable visual concepts, we train a task conditional visual classifier, whose task is defined by a task feature.
The classifier is used as an answering unit, where a task feature is inferred from a question.
To train the task conditional visual classifier without task annotations, we propose an unsupervised task discovery technique based on linguistic knowledge sources such as structured lexical databases, \eg, WordNet~\cite{fellbaum1998wordnet} and region descriptions~\cite{krishna2017visual}.
We claim that the proposed transfer learning framework helps generalization in VQA with out-of-vocabulary answers.
The main contribution of our paper is three-fold:
\vspace{-0.1cm}
\begin{itemize}
\item{We present a novel transfer learning algorithm for visual question answering based on a task conditional visual classifier.} \vspace{-0.1cm}
\item{We propose an unsupervised task discovery technique for learning task conditional visual classifiers without explicit task annotations.}  \vspace{-0.1cm}
\item{We show that the proposed approach handle out-of-vocabulary answers through knowledge transfer from visual dataset without question annotations.}
\vspace{-0.1cm}
\end{itemize}

The rest of the paper is organized as follows. Section~\ref{sec:related_works} discusses prior works related to our approach.
We describe the overall transfer learning framework in Section~\ref{sec:algorithm_overview}.
Learning visual concepts by unsupervised task discovery is described in Section~\ref{sec:task_discovery}.
Section~\ref{sec:experiments} analyzes experimental results and Section~\ref{sec:conclusion} makes our conclusion.

\begin{figure*}[!t]
	\centering
	\includegraphics[width=0.98\linewidth] {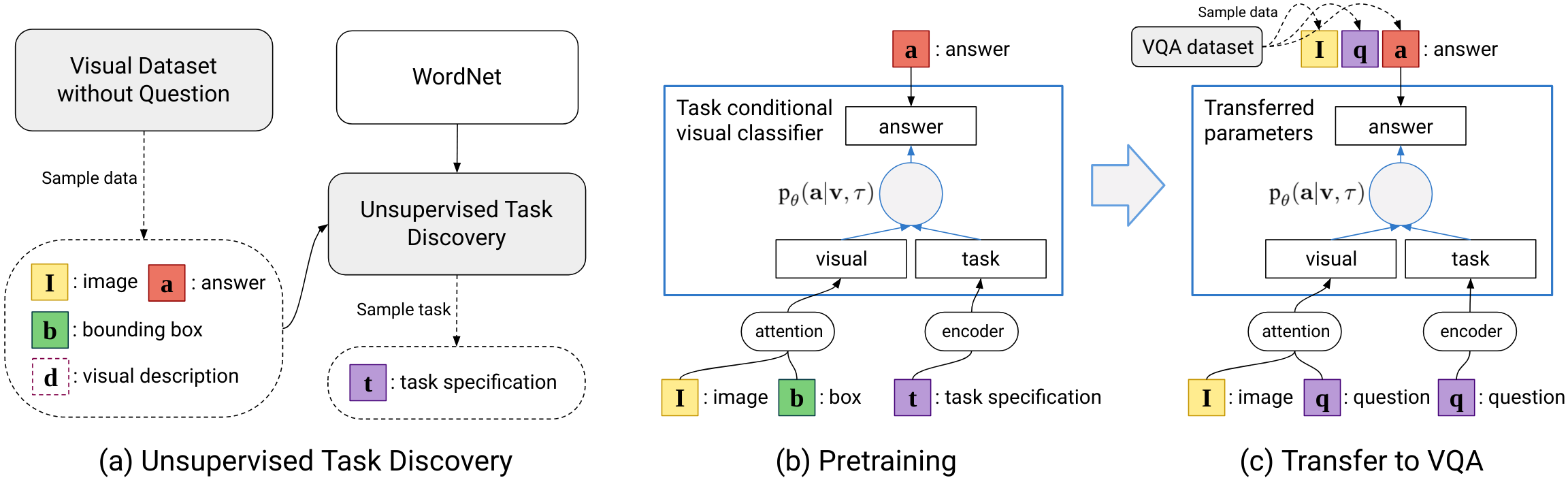}
	\vspace{0.05cm}
	\caption{
		\textbf{Overview of the proposed algorithm.}
		(a) Unsupervised task discovery samples a task specification for a sampled visual data $(\vara, \varI, \varb)$, where $\varI$, $\varb$ and $\vara$ are an image, a bounding box and an label (answer), respectively.
		It leverages linguistic knowledge sources such as visual description $\vard$ and WordNet.
		(b) A visual data with a task specification, denoted by $(\vara, \varI, \varb, \vart)$, is employed to pretrain a task conditional visual classifier.
		(c) The pretrained task conditional visual classifier is transferred to VQA and the parameters are frozen. Attention layer and question encoder are learned from scratch with VQA dataset. The terms label and answer are used interchangeably.
	}
	\label{fig:overview}
	\vspace{-0.3cm}
\end{figure*}

\section{Related Works}
\label{sec:related_works}

The standard VQA evaluation assumes identically distributed train and test set~\cite{antol2015vqa, malinowski2014multi, zhu2016visual7w}.
As this evaluation setting turns out to be vulnerable to models exploiting biases in training set~\cite{goyal2017making}, several alternatives have been proposed. 
One approach is to reduce observed biases either by balancing answers for individual questions~\cite{goyal2017making} or by providing different biases to train and test sets intentionally~\cite{agrawal2017don}.
Another approach is to construct compositional generalization split~\cite{agrawal2017c, johnson2017clevr} whose question and answer pairs in test set are formed by novel compositions of visual concepts and question types appearing in the training set.
This split is constructed by repurposing an existing VQA dataset~\cite{agrawal2017c} or by constructing a synthetic dataset~\cite{johnson2017clevr}. 
The problem setting studied in this paper is similar to \cite{kafle2018dvqa, teney2016zero} in the sense that out-of-vocabulary answers are used for testing, but unlike the prior work, we formulate the problem as a transfer learning, where out-of-vocabulary answers are learned from  external visual data.

External data are often employed in VQA for better generalization.
Convolutional neural networks~\cite{he2016deep, krizhevsky2012imagenet} pretrained on ImageNet~\cite{Imagenet} is a widely accepted standard for diverse VQA models~\cite{fukui2016multimodal, yang2016stacked}.
As an alternative, object detector~\cite{ren2015faster} trained on the Visual Genome dataset~\cite{krishna2017visual} is employed to extract pretrained visual features~\cite{anderson2017updown}.
Pretrained language models such as word embeddings~\cite{pennington2014glove} or sentence embeddings~\cite{kiros2015skip} are frequently used to initialize parameters of question encoders~\cite{fukui2016multimodal, noh2016image, teney2017tips}. 
Exploiting information retrieval from knowledge base~\cite{auer2007dbpedia,bollacker2008freebase} or external vision algorithms~\cite{wang2017vqa} to provide additional inputs to VQA models are investigated in~\cite{wang2017fvqa, wang2017vqa, wu2016ask}.
Transfer between VQA datasets is studied in~\cite{hu2018learning}.
Sharing aligned image-word representations between VQA models and image classifiers has been proposed in~\cite{gupta2017aligned} to exploit external visual data.

Transfer learning from external data to cope with out-of-vocabulary words has hardly been studied in VQA, but is actively investigated in novel object captioning~\cite{hendricks2016deep, lu2018neural, venugopalan2017captioning, yao2017incorporating}.
For example, \cite{hendricks2016deep} and \cite{venugopalan2017captioning} decompose image captioning task into visual classification and language modeling, and exploit unpaired visual and linguistic data as additional resources to train visual classifier and language model, respectively.
Recent approaches incorporate pointer networks~\cite{vinyals2015pointer} and learn to point an index of word candidates~\cite{yao2017incorporating} or an associated region~\cite{lu2018neural}, where the word candidates are detected by a multi-label classifier~\cite{yao2017incorporating} or an object detector~\cite{lu2018neural} trained with external visual data.
However, these algorithms are not directly applicable to our problem setting because they focus on predicting object words without task specification while the task conditional visual recognition is required for VQA.

Our problem setting is closely related to zero-shot learning~\cite{frome2013devise, huang2015learning, lampert2014attribute, larochelle2008zero, xian2017zero}, where out of vocabulary answers are considered in classification. 
Zero-shot learning aims to recognize objects or classes that are unseen during training.
As it aims generalization to completely unseen classes, any exposure to zero-shot classes during training is strictly prohibited~\cite{xian2017zero}.
On the contrary, our goal is to exploit class labels available in external dataset.

\section{Transfer Learning Framework}
\label{sec:algorithm_overview}

The main goal of our work is to handle out-of-vocabulary answers in VQA by learning visual concepts from off-the-shelf visual dataset and transfering the concepts to VQA for answering the questions.
Inspired by the fact that VQA can be thought of as a task conditional classification problem, where tasks are defined by questions, we introduce a task conditional visual classifier, which generates an answer from a visual input and a task specification, as a medium for learning and transfering visual concepts.
Figure~\ref{fig:overview} illustrates the overall framework of the proposed approach.
We pretrain the task conditional visual classifier using visual dataset without questions or task specifications via unsupervised task discovery, and adapt it to VQA models by transferring the learned parameters.
We describe the task conditional visual classifier and how it is pretrained and transferred to VQA in the rest of this section.


\subsection{Task conditional visual classifier}
\label{subsec:taskcondtionalclassifier}

Task conditional visual classifier is a function taking a visual feature $\varv \in \real^d$ and a task feature $\tau \in \real^k$ and producing a probability distribution of answers or labels $\vara\in [0, 1]^{l}$, where the terms answer and label are used interchangeably based on context hereafter.
The classifier formulated as a neural network with parameter $\theta$ models a conditional distribution $\prob_\theta(\vara | \varv, \tau)$.
Note that two inputs $\varv$ and $\tau$ are typically obtained by encoders $\varv_\phi(\cdot)$ and $\tau_\eta(\cdot)$.

In the proposed transfer learning scenario, a task conditional visual classifier is pretrained with off-the-shelf visual dataset, \eg, Visual Genome~\cite{krishna2017visual}, and transfered to VQA.
In the pretraining stage, the parameters of the classifier and two feature encoders $\theta$, $\phi_\text{pre}$ and $\eta_\text{pre}$ are jointly learned.
This stage allows the task conditional visual classifier to handle diverse visual recognition tasks by learning the task feature $\tau$.
Transfer learning to VQA is achieved by reusing parameter $\theta$ and adapting new encoders $\varv_{\phi_\vqa}(\cdot)$ and $\tau_{\eta_\vqa}(\cdot)$ to the learned task conditional visual classifier.

\begin{figure}[t]
	\centering
	\includegraphics[width=0.98\linewidth] {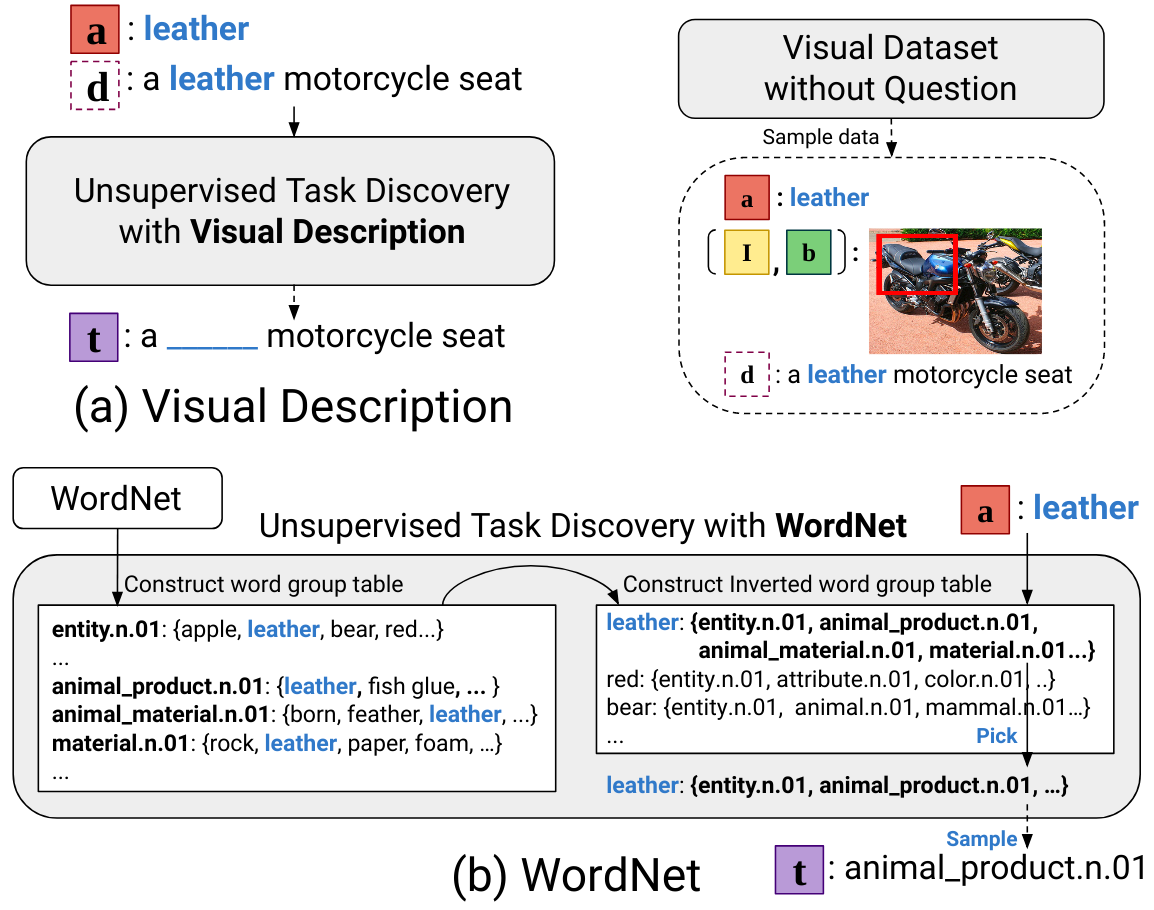}
	\vspace{0.05cm}
	\caption{
		\textbf{Unsupervised task discovery with two different linguistic knowledge sources.} (a) For visual description, task specification is generated by replacing a visual word (label) from the description into a blank. (b) For WordNet, task specification is a synset from one of the labels hypernyms. We use {\em inverted wordset} to sample a synset from an input label.
		See Section~\ref{sec:task_discovery} for details.
	}
	\label{fig:unsupervised_task_discovery}
	\vspace{-0.2cm}
\end{figure}

\begin{figure*}[t]
	\centering{

	\begin{minipage}[t]{0.245\linewidth}
	\includegraphics[width=0.99\linewidth] {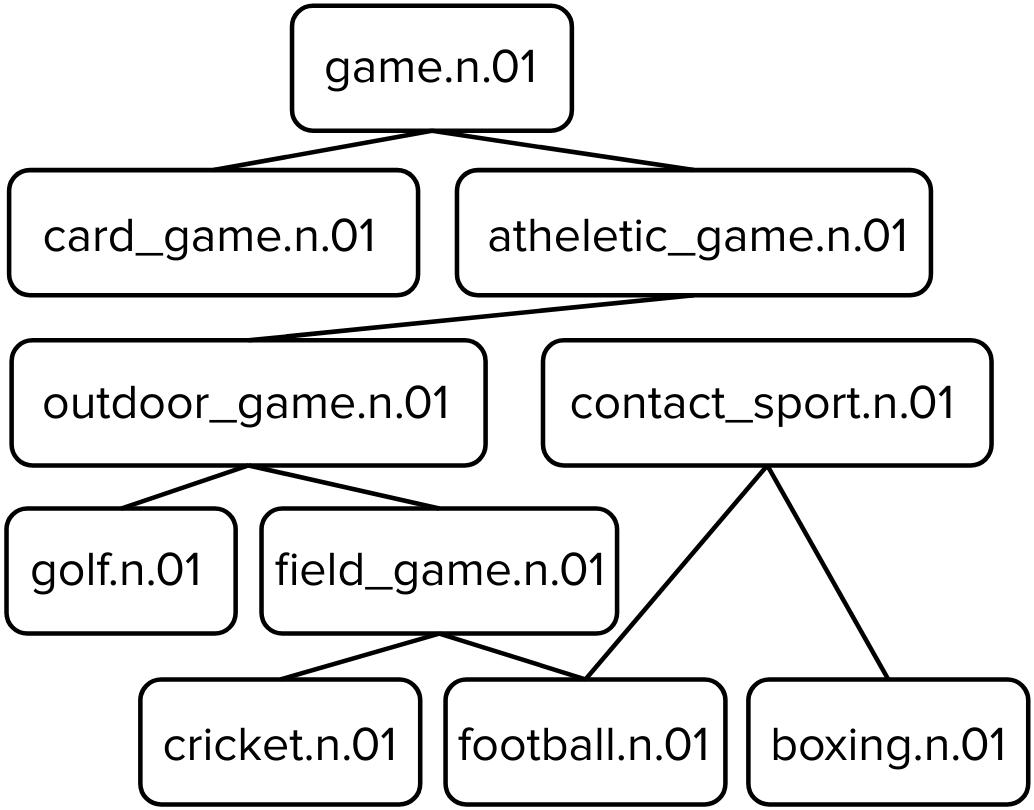}
	\end{minipage}
	\begin{minipage}[t]{0.72\linewidth}
	\raisebox{0.9\height}{
	\small
	\begin{tabular}
		{
			 @{}C{2.2cm}@{} |	@{}L{10.4cm}@{} 
		}    
	    \hline    
		task $\vart_w$ & word groups \vspace{0.02cm} \\
		\hline
	    \vspace{0.02cm}
	    appliance.n.02 & freezer, hair dryer, refrigerator, fridge, oven, dishwasher, washing machine, ... \vspace{0.02cm}\\
	    opening.n.10 & manhole, rear window, exit, nozzle, car window, coin slot, spout, vent, ... \vspace{0.02cm}\\
	    food.n.02 & potatoes, french fries, chicken, melted cheese, tomatoes, sausage, vegetable, ... \vspace{0.02cm}\\
	    move.v.03 & twisted, ruffled, curving, curved, wheeled, rolled, coiled, turned, rippled, ... \vspace{0.02cm}\\
	    act.n.02 & fishing, skateboarding, sailing, playing baseball, surfing, traveling, driving, ... \vspace{0.02cm}\\
	    area.n.01 & middle, in corner, center, parking space, playground, landscape, neighborhood, ...\vspace{0.02cm}\\
	    color.n.01 & yellow, pink, red, beige, royal blue, amber, aqua, dark red, olive green, teal, ... \vspace{0.02cm}\\
	    \hline
	\end{tabular}
	}
	\end{minipage}
	}
	\vspace{0.1cm}
	\caption{
	\textbf{Illustration of WordNet and constructed word group table.}
	(Left) A subgraph of the WordNet~\cite{fellbaum1998wordnet}.
	Complex hierarchy of words reveals the diverse categorization of each words.
	(Right) A set of words sharing common parents in the tree is grouped as a single word group. Diverse grouping of words reveals diverse visual recognition tasks that can be defined on each word group.
	}
	\vspace{-0.3cm}
	\label{fig:wordset_examples}
\end{figure*}

\subsection{Pretraining}
\label{subsec:pretraining}

Learning the task conditional visual classifier is naturally formulated as the problem to maximize the following expected log likelihood:
\begin{equation}
\theta^*, \phi^*_\pre, \eta^*_\pre = 
\argmax_{\theta, \phi_\pre, \eta_\pre} 
\mathbb{E}_{\prob_\mathcal{D}}
\left[ \text{log} \, \prob_\theta(\vara | 
\varv_{\phi_\pre}(\varI, \varb),
\tau_{\eta_\pre}(\vart)) \right],
\label{eq:pretrain}
\end{equation}
where $\varv_{\phi_\pre}(\varI, \varb)$ is a visual feature based on an image $\varI$ and a bounding box $\varb$, and $\tau_{\eta_\pre}(\vart)$ is a task feature encoded from a task specification $\vart$, $\vara$ is an answer sampled from data distribution and it satisfies $\vara \in \mathcal{A}$, and $\{ \theta, \phi_\pre, \eta_\pre \}$ are model parameters.
We obtain $\varv_{\phi_\pre}(\varI, \varb)$ using a learnable attention network parametrized by $\phi_\text{pre}$ on top of off-the-shelf feature extractor~\cite{anderson2017updown}, where the bounding box position $\varb$ is used as a key to the attention.
The optimization in Eq.~\eqref{eq:pretrain} requires a joint distribution, $\prob_\mathcal{D}(\vara, \varI, \varb, \vart)$, which is not accessible in the external datasets in our setting due to missing task specifications $\vart$.
Section~\ref{sec:task_discovery} describes how to model the joint distribution $\prob_\mathcal{D}(\vara, \varI, \varb, \vart)$ with the visual annotations and linguistic knowledge sources.

\subsection{Transfer learning for VQA}
\label{subsec:transfer_vqa}

As illustrated in Figure~\ref{fig:overview}, the proposed VQA model contains a task conditional visual classifier $\prob_\theta(\vara | \varv, \tau)$. 
The pretrained visual concepts are transferred to VQA by sharing the learned parameters $\theta$.
Then, learning a VQA model is now formulated as learning input representations $\varv$ and $\tau$ for $\prob_\theta(\vara| \varv, \tau)$, which is given by
\begin{equation}
\phi_\vqa^*, \eta_\vqa^* = \argmax_{\phi_\vqa, \eta_\vqa} \mathbb{E}_{\prob_\vqa} \left[ \log \, \prob_\theta(\vara|\varv_{\phi_\vqa}(\varI, \varq), \tau_{\eta_\vqa}(\varq)) \right],
\label{eq:transfer}
\end{equation}
where $\varv_{\phi_\vqa}(\varI, \varq)$ is an encoded visual feature with an image $\varI$ and a question $\varq$ using an attention mechanism with parameter $\phi_\vqa$ and a off-the-shelf feature extractor~\cite{anderson2017updown}. 
A task feature $\tau_{\eta_\vqa}(\varq)$ encodes a question $\varq$ using parameter $\eta_\vqa$.
The joint distribution of a training dataset for VQA, $\prob_\text{vqa}(\vara, \varI, \varq)$, is required for optimization, where answers from the distribution satisfy $\vara \in \mathcal{A} - \mathcal{B}$.
We learn $\phi_\vqa$ and $\eta_\vqa$ by maximizing the likelihood of the objective in Eq.~\eqref{eq:transfer} while the parameter for the pretrained task conditional visual classifier $\theta$ remains fixed.

\vspace{-0.3cm}
\paragraph{Weakly supervised task regression}
Utilizing a pretrained task conditional visual classifier for the visual recognition specified by a question $\varq$ requires to infer an optimal task feature $\tau^*_\varq$.
This requirement introduces a learning problem---task regression---that optimizes an encoder $\tau_{\eta_\vqa}(\varq)$ to predict $\tau^*_\varq$ correctly.
Because directly minimizing error $\mathcal{E}(\tau^*_\varq, \tau_{\eta_\vqa}(\varq))$ requires additional supervision about the tasks, we instead exploit VQA data as a source of weak supervision.
We optimize indirect loss $-\mathbb{E}_{\tau^*_\varq(\vara|\varv)}[\log \, \prob_\theta(\vara|\varv, \tau_{\eta_\vqa}(\varq))]$, which encourages the answer distributions conditioned on $\tau^*_\varq$ and $\tau_{\eta_\vqa}(\varq)$ to be similar.
By assuming that true task conditional answer distribution $\tau^*_\varq(\vara|\varv)$ is implicitly modeled in VQA dataset, we employ Eq.~\eqref{eq:transfer} as the objective function for weakly supervised task regression.

\vspace{-0.3cm}
\paragraph{Out-of-vocabulary answering}
We learn a VQA model by adapting input representations while fixing the pretrained task conditional visual classifier $\prob_\theta(\vara|\varv, \tau)$.
This strategy allows a model to focus on learning to infer a visual recognition task $\tau_{\eta_\vqa}(\varq)$ from questions, which does not require data for all possible answers.
Once the task feature $\tau$ is inferred, the learned task conditional visual classifier $\prob_\theta(\vara|\varv, \tau)$ can answer the pretrained visual concepts including out-of-vocabulary ones.

\vspace{-0.3cm}
\paragraph{Matching visual features}
To reuse pretrained visual classifier in VQA without fine-tuning, semantics of visual features $\varv$ should not change by learning with VQA dataset.
This is fulfilled in recent approaches for VQA models that do not fine-tune pretrained visual feature extractors and focus on learning attention mechanism~\cite{kim2016hadamard} over the extracted feature map.
In our setting, we simply use the identical visual feature extractor~\cite{anderson2017updown} for both pretraining and VQA.

\section{Unsupervised Task Discovery}
\label{sec:task_discovery}

Learning a task conditional visual classifier with the off-the-shelf visual dataset~\cite{krishna2017visual} is not straightforward due to missing annotation for task specifications, which is necessary to learn an encoder for a task specification vector $\tau$.
To address this issue, we propose unsupervised task discovery which samples a task specification $\vart$ from a task distribution modeled by exploiting linguistic knowledge sources.

\subsection{Leveraging linguistic knowledge sources}
\label{subsec:linguistic_sources}

A visual recognition task given by a question typically defines a mapping from visual inputs to a set of possible visual concepts, that is, a word group.
For example, a question of ``what is the woman holding?" defines a visual recognition task finding \textit{holdable objects} in an image, which is a classification over a word group $\{\text{ball, racket, cup, ...} \}$.
This intuition leads to a simple approach for modeling distribution of task description $\vart$ by treating a task as a word group (\ie \textit{holdable objects}).
The main reason to use linguistic knowledge sources for the unsupervised task discovery is that the word groups are often accessible in the linguistic knowledge sources.
We consider two linguistic knowledge sources: 1) visual descriptions provided with a visual data and 2) a structured lexical database called WordNet~\cite{fellbaum1998wordnet}.
Figure~\ref{fig:unsupervised_task_discovery} illustrates the overview of our approach.

\subsection{Visual description}
\label{subsec:visual_description}
We use Visual Genome~\cite{krishna2017visual} as an off-the-shelf visual dataset, which determines a data distribution $\prob_\mathcal{V}(\vara, \varI, \varb, \vard)$ based on a set of quadruples $(\vara, \varI, \varb, \vard)$ including visual descriptions $\vard$.
The description in this dataset is designed to mention answer $\vara$ explicitly, so that relation between the answer and the description is clear.

To this end, we define task specification $\vart_d$ by replacing the answer in visual description to a special word $<$blank$>$, which is formally denoted as $\vart_d=\rho(\vard, \vara)$, where $\rho(\vard, \vara)$ is a function generating a blanked description.
The subscript in $\vart_d$ means that a task specification is extracted based on a visual description.
Based on this definition, joint distribution, $\prob_\mathcal{D}(\vara, \varI, \varb, \vart_d) = \int \prob(\vara, \varI, \varb, \vart_d, \vard) d\vard$, with task specification is given by
\begin{align}
\prob(\vara, \varI, \varb, \vart_d, \vard) = \prob(\vart_d|\vard, \vara)\prob_\mathcal{V}(\vara, \varI, \varb, \vard) 
\label{eq:sample_description}
\end{align}
where $\prob(\vart_d|\vard, \vara) = \delta\big(\vart_d, \rho(\vard, \vara)\big)$ is a delta function that returns $1$ if two inputs are identical and $0$ otherwise.
As illustrated in Figure~\ref{fig:unsupervised_task_discovery}a, we sample data required for pretraining $(\vara, \varI, \varb, \vart_w)$ by first sampling a visual data $(\vara, \varI, \varb, \vard)$ and then sampling a task specification $\vart_d$ from $\prob(\vart_d|\vard,\vara)$.
This procedure results in sampling description $\vard$ as well, but we do not care about it when we pretrain the task conditional visual classifier.
For pretraining, we encode $\vart_d$ into a task feature $\tau_{\eta_\pre}(\vart_d)$ based on a gated recurrent unit~\cite{chung2014empirical} because $\vart_d$ is a sequence of words.

The main reason to use a blanked description for a task specification is that it is effective to define a set of candidate words.
For example, a blanked description ``\textit{a man is holding \underline{\hspace{0.8cm}}}" restricts  candidate words for the blank to a set of objects that can be held.
Therefore, a blanked description can be used to determine a word group implicitly that represents a visual recognition task.

\begin{figure*}[!t]
	\centering
	\includegraphics[width=0.95\linewidth] {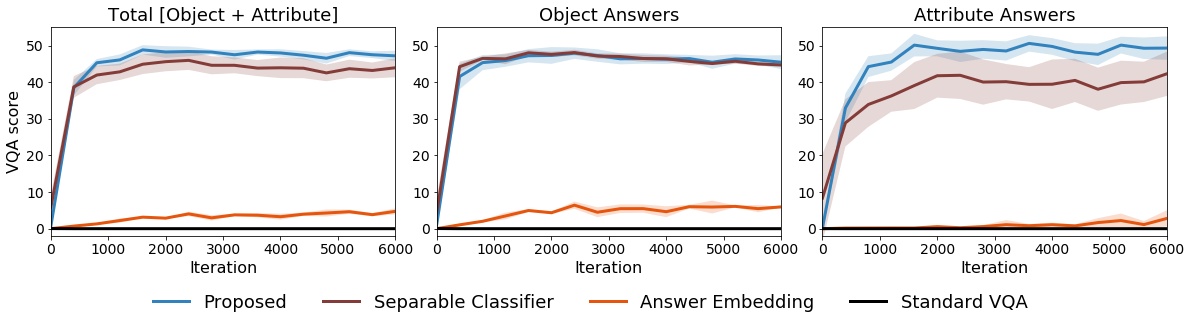}
	\vspace{-0.2cm}
	\caption{
		\textbf{Model comparisons.}
		Exploiting external data with the unsupervised task discovery boosts performance of the proposed model and separable classifier significantly while the separable classifier shows limited gains on attribute answers with large variations.
		}
	\label{fig:vqa_model_comparison}
\end{figure*}

\subsection{WordNet}
\label{subsec:wordnet}
WordNet~\cite{fellbaum1998wordnet} is a lexical database represented with a directed acyclic graph of disambiguated word entities, called synsets.
A sample subgraph of WordNet is illustrated in Figure~\ref{fig:wordset_examples} (left).
The graph represents a hierarchical structure of words, where the parents of a node correspond to hypernyms of the word in the child.
In WordNet, we define a task specification $\vart_w$ as a synset for a node that is a common ancestor of multiple words because a set of words sharing a common ancestor constructs a word group and the word group may also define a visual recognition task.

A procedure for sampling a task specification $\vart_w$ based on WordNet and a visual data $(\vara, \varI, \varb, \vard)$ is illustrated in Figure~\ref{fig:unsupervised_task_discovery}b.
The main idea for the procedure is to model a task distribution conditioned on an answer $\prob(\vart_w|\vara)$ as a uniform distribution over possible word groups that the answer belongs to, where a task specification $\vart_w$ is a common ancestor of words in an word group.
Modeling the distribution $\prob(\vart_w|\vara)$ requires two stages: 1) constructing a word group table, which maps a task specification to a word group and 2) constructing an inverted word group table, which maps an answer word to a set of task specifications.
The inverted word group table is used to retrieve a set of possible task specifications for an answer $\vara$ and the distribution $\prob(\vart_w|\vara)$ is the uniform distribution over task specifications in the set. 
Given the distribution $\prob(\vart_w|\vara)$, the joint distribution, $\prob_\mathcal{D}(\vara, \varI, \varb, \vart_w) = \int \prob(\vara, \varI, \varb, \vart_w, \vard) d\vard$, is given by
\begin{equation}
\prob(\vara, \varI, \varb, \vart_w, \vard) = \prob(\vart_w|\vara)\prob_\mathcal{V}(\vara,\varI,\varb, \vard).
\end{equation}
Therefore we sample a quadruple $(\vara, \varI, \varb, \vard)$ from the visual dataset and sample a task specification $\vart_w$ subsequently. 
While this procedure samples a description as well, we marginalize it out. 
For pretraining, we encode $\vart_w$ into a task specification vector $\tau_{\eta_\pre}(\vart_w)$ based on a word embedding function that is learned from scratch.

The word group table is constructed by selecting a synset of a node in WordNet as a task specification $\vart_w$ and mapping it to a set of words (a word group) corresponding to all its descendants. 
Any word group can be defined regardless of its level in WordNet hierarchy and the part-of-speech of its members; the biggest word group contains all words in WordNet and its task specification corresponds to the root of WordNet.
We illustrate the constructed word group table in Figure~\ref{fig:wordset_examples}.
The inverted word group table is constructed in a similar way to an inverted index of the word group table, but the range of mapping is not a set of indices but a set of task specifications.

\begin{figure*}[!t]
\centering
\includegraphics[width=0.95\linewidth]{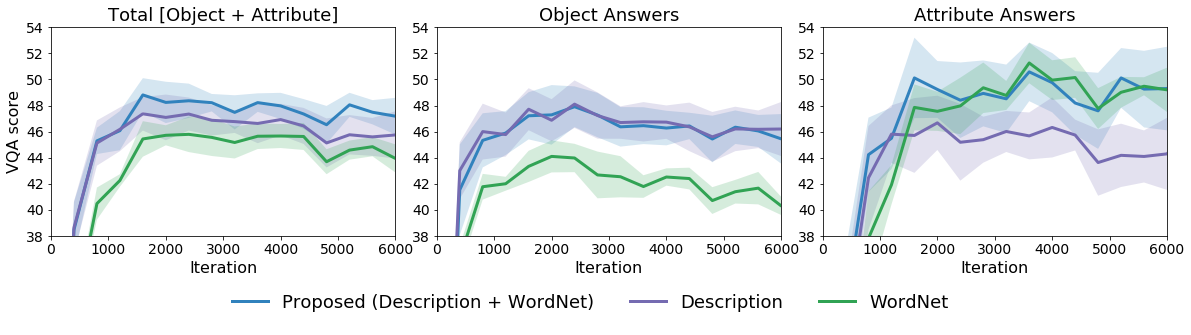}
\vspace{-0.2cm}
\caption{
\textbf{Data comparisons.}
Using visual description and WordNet shows different generalization characteristics and combining them brings additional improvement.}
\label{fig:vqa_data_comparison}
\vspace{-0.05cm}
\end{figure*}

\begin{figure}[!t]
\centering
\includegraphics[width=1\linewidth]{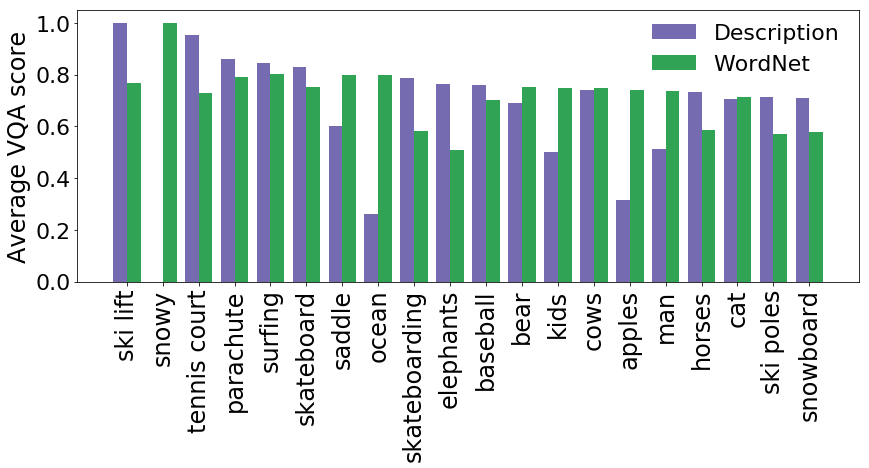}
\caption{
\textbf{Complementary characteristics of data.}
Visual description and WordNet show complementary characteriscs in terms of VQA score for different answers.
}
\label{fig:score_per_answer}
\vspace{-0.2cm}
\end{figure}

\section{Experiments}
\label{sec:experiments}

We evaluate how effectively the proposed framework leverages the external data without questions to answer out-of-vocabulary words in visual question answering.
We compare the proposed method with the baselines equipped with idea for zero-shot image classification~\cite{frome2013devise} and novel object captioning~\cite{hendricks2016deep, venugopalan2017captioning}, which are related to the proposed problem.
We also analyze the impact of the external data used for pretraining, and visualize the mapping between questions and task specifications learned by weakly supervised task regression.
Notably, the evaluation setting is not comparable to that from zero-shot learning literature~\cite{xian2017zero}, as our objective is not generalization to completely unseen classes, but exploiting class labels exposed in the external data.
We open-sourced all the codes and datasets used in the experiments to facilitate reproducing the result in this paper~\footnote{\footnotesize \url{https://github.com/HyeonwooNoh/vqa\_task\_discovery}}.

\subsection{Datasets}

\paragraph{Pretraining}

We learn visual concepts of most frequently observed 3,000 objects and 1,000 attributes in the Visual Genome dataset~\cite{krishna2017visual}. 
To pretrain the task conditional visual classifier, we construct external visual data with bounding box annotations, which are provided with region descriptions.
Then, visual words (answers) are extracted from region descriptions to construct visual data quadruples $(\vara, \varI, \varb, \vard)$.
We use 1,169,708 regions from 80,602 images to construct the training data.
To use WordNet~\cite{fellbaum1998wordnet}, we map visual words to synsets using synset annotations from Visual Genome dataset, and the words that are not covered by the annotations are mapped to synsets using Textblob~\cite{textblob}. 

\begin{figure*}[!t]
\centering
\includegraphics[width=0.95\linewidth] {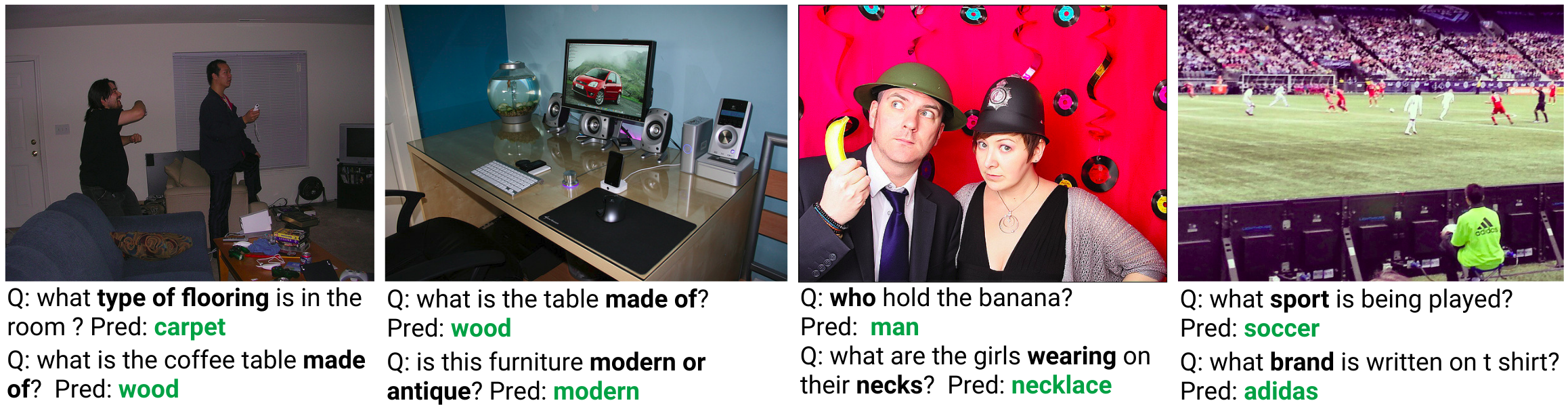}
\vspace{0.15cm}
\caption{
\textbf{Out-of-vocabulary answers with diverse types of concepts.}
Green color denotes correct answers. All predicted answers are from out-of-vocabulary answers. The proposed model successfully predicts diverse out-of-vocabulary answers depending on questions.
}
\label{fig:qualitative}
\end{figure*}
%
\begin{table*}[!t]\small
\centering
\caption{
\textbf{Result of weakly supervised task discovery.}
We retrieve questions for each task specification based on the similarity score with task features.
Results show that appropriate task specifications are regressed from each question.
Note that no explicit supervision is used for learning the mapping between questions and task specifications.
}
\vspace{0.15cm}
\scalebox{1.0}{
    \begin{tabular}
    	{
    		 @{}L{3.0cm}@{}| @{}L{13.3cm}@{}    
    	}    
        \hline    
        Task specification $\vart_w$ & Questions \vspace{0.02cm} \\
        \hline
        organic\_process.n.01 & what are the giraffes doing? / what are the animals doing? / what are the giraffes doing in this picture?\\
        athletic\_game.n.01 & what type of sport ball is shown? / what type of sport are the men participating in? \\
        furniture.n.01 & what piece of furniture are the cats sitting on? / what furniture is the cat sitting on?\\
        fruit.n.01 & what type of fruit is the animal eating? / what type of fruit juice is on the counter?\\
        time\_period.n.01 & what kind of season is it? / what type of season is it? / what holiday season is it? / which season is it?\\
        tool.n.01 & what utensil is in the person 's hand? / what utensil is laying next to the bread? \\
        hair.n.01 & what hairstyle does the surfer have ? / what type of hairstyle does this man have ? \\
        \hline
    \end{tabular}
}
\label{tab:task_regression} 
\vspace{-0.2cm}
\end{table*}

\vspace{-0.3cm}
\paragraph{Dataset construction}
We repurpose VQA v2 dataset to construct a training/test split as illustrated in Figure~\ref{fig:zeroshot_setting}.
We use training and validation set of VQA v2.
To ensure that every out-of-vocabulary answer appears during pretraining, we randomly select out-of-vocabulary answers from all the visual words used for pretraining (954 out of 3,813).
Since we focus on transferability of visual words, answers about yes/no and numbers are not considered in our evaluation.
Based on the selected out-of-vocabulary answers, we generate 3 questions splits---462,788 for training, 51,421 for validation, and 20,802 for testing.
The training and validation splits do not contain out-of-vocabulary answers while the test split consist of out-of-vocabulary answers only.
For evaluation of models, we follow the standard VQA protocol with $10$ ground-truth answers for each question~\cite{antol2015vqa}.

\subsection{Baselines}
Since leveraging external visual data for visual question answering with out-of-vocabulary answers has hardly been explored, there is no proper evaluation benchmark and we employ the following baselines to compare with our algorithm:
1) \textit{answer embedding}, which employs idea from zero-shot image classification~\cite{frome2013devise} that learns mapping from visual features to pretrained answer embedding, where we use GloVe~\cite{pennington2014glove} to embed each answer, and
2) \textit{separable classifier}, which adopts idea from novel object captioning~\cite{hendricks2016deep, venugopalan2017captioning} that learns visual and language classifier separately and combines them by element-wise sum of logits for joint inference.
Note that the separable classifier and our proposed model are trained with the same data.
%
\begin{figure*}[!t]
	\centering
	\includegraphics[width=0.95\linewidth] {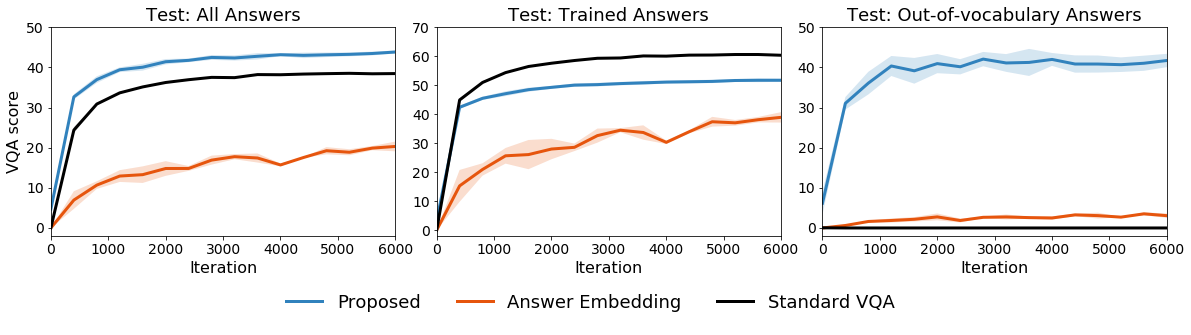}
	\vspace{-0.1cm}
	\caption{
		\textbf{Combining knowledge from VQA and external visual data.}
		Evaluation results on a test set containing both out-of-vocabulary answers and trained answers.
		The proposed model shows relatively lower performance on trained answers but significantly better performance on out-of-vocabulary answers.
		In total, the proposed model shows the best performance.}
	\label{supple_fig:vqa_all}
	\vspace{-0.2cm}
\end{figure*}

\subsection{Results}

\paragraph{Model comparisons}
Figure~\ref{fig:vqa_model_comparison} illustrates model comparison results with the baselines.
For this experiment, we perform VQA adaptation with 6 different random seeds and plot their mean and standard deviation.
The standard VQA model fails to predict any out-of-vocabulary answers (\ie, 0 VQA score) because there is no clue for inferring out-of-vocabulary answers.
Answer embedding baseline generalizes slightly better by exploiting the similarity of answer words in the embedding space, but the improvement is marginal.
%
Using off-the-shelf visual data and task specifications from linguistic knowledge sources dramatically improves performance both on the separable classifier and the proposed model.
However, independent consideration of visual data and task specifications in separable classifier has a critical limitation of modeling joint interaction between task specifications and visual features.
Especially, this baseline shows substantially lower performance on attribute answers, which have significant variations depending on tasks.
Note that the bias in the VQA training set cannot be exploited in the proposed evaluation setting, as the evaluation is performed with out-of-distribution answers only.

\vspace{-0.3cm}
\paragraph{Data comparisons}

Figure~\ref{fig:vqa_data_comparison} illustrates the effect of different linguistic sources in our algorithm.
Two models learned using visual description and WordNet, respectively, have complementary characteristics and additional improvement is achieved by exploiting both data.
More detailed complementary characteristics of models are illustrated in Figure~\ref{fig:score_per_answer}, where we visualize average VQA scores for 20 answers.

\vspace{-0.3cm}
\paragraph{Qualitative results}
Figure~\ref{fig:qualitative} shows examples of predicted answers from the proposed model.
The proposed model correctly predicts out-of-vocabulary answers for questions asking diverse visual concepts such as type of flooring, material, type of sport and brand.

\vspace{-0.3cm}
\paragraph{Weakly supervised task regression}
Given that task specifications extracted from WordNet models diverse visual recognition tasks, matching them to relevant questions is useful for categorization of VQA data and model interpretation.
As we learn VQA models by task regression, this matching can be performed by comparing the encoded task feature from a question $\tau_{\eta_\vqa}(\varq)$ and the vector from a task specification $\tau_{\eta_\pre}(\vart_w)$. 
For each $\tau_{\eta_\pre}(\vart_w)$, we sorted questions in a descending order of dot product similarity between $\tau_{\eta_\pre}(\vart_w)$ and $\tau_{\eta_\vqa}(\varq)$. 
In the sorted question list, the most similar questions are visualized in Table~\ref{tab:task_regression}.
The visualization shows that the weakly supervised task regression successfully trains a question encoder that match a question to a relevant task feature.

\subsection{Combining knowledge learned by VQA}

While we focus on learning visual concepts from external visual data, VQA dataset is still a valuable source of learning diverse knowledges.
Especially, some answers are not visual words and require visual reasoning.
For example, yes and no are one of the most frequent answers in the VQA dataset~\cite{antol2015vqa} but it is not straightforward to learn these answers only with the external visual data.
Therefore, we consider combining knowledge learned from VQA dataset and from external visual data.

We construct a split of the VQA dataset consisting of 405,228 training, 37,031 validation, and 172,681 test questions.
The training and validation set do not contain any out-of-vocabulary answers and test set contains out-of-vocabulary answers.
However, contrary to the main experiment, the test set also contains training answers which include logical answers, numbers and visual words.
The list of out-of-vocabulary answers are identical to that of the main experiment.
Among 172,681 test questions, 103,013 questions can be answered only with the training answers.

To combine knowledges from VQA dataset and external visual data, we learn a VQA model with two task conditional visual classifiers; 
we fine-tune one classifier for adapting answers requiring visual reasoning (\ie, numbers and yes/no) and fix the other classifier for visual answers including out-of-vocabulary answers.
After training the VQA model, we combine two logits by element-wise sum and pick the answer with the highest score in the inference.

The results are presented in Figure~\ref{supple_fig:vqa_all}.
Models in each method are trained with 6 different random seeds and their mean and standard deviation are plotted.
Overall, the proposed model performs the best.
While the standard VQA model achieves the best performance for training answers, it fails to predict any out-of-vocabulary answers.
The answer embedding baseline somewhat generalizes to out-of-vocabulary answers, but constraints in the answer embedding degrade its performance on answers in training set.


\section{Conclusion}
\label{sec:conclusion}
We present a transfer learning approach for visual question answering with out-of-vocabulary answers.
We pretrain a task conditional visual classifier with off-the-shelf visual and linguistic data based on unsupervised task discovery.
The pretrained task conditional visual classifier is transferred to VQA adaptively.
The experimental results show that exploiting external visual and linguistic data boosts performance in the proposed setting and training with unsupervised task discovery is important to model interaction between visual features and task specifications.


\vspace{-0.2cm}
\paragraph{Acknowledgments}
This research was partly supported by Kakao and Kakao Brain and Korean ICT R\&D program of the MSIP/IITP grant [2016-0-00563, 2017-0-01778].

\bibliography{vqa_transfer}

\begin{thebibliography}{10}\itemsep=-1pt

\bibitem{agrawal2017don}
Aishwarya Agrawal, Dhruv Batra, Devi Parikh, and Aniruddha Kembhavi.
\newblock {Don't Just Assume; Look and Answer: Overcoming Priors for Visual
  Question Answering}.
\newblock In {\em CVPR}, 2018.

\bibitem{agrawal2017c}
Aishwarya Agrawal, Aniruddha Kembhavi, Dhruv Batra, and Devi Parikh.
\newblock {C-VQA: A Compositional Split of the Visual Question Answering (VQA)
  v1. 0 Dataset}.
\newblock {\em arXiv preprint arXiv:1704.08243}, 2017.

\bibitem{anderson2017updown}
Peter Anderson, Xiaodong He, Chris Buehler, Damien Teney, Mark Johnson, Stephen
  Gould, and Lei Zhang.
\newblock {Bottom-Up and Top-Down Attention for Image Captioning and Visual
  Question Answering}.
\newblock In {\em CVPR}, 2018.

\bibitem{hendricks2016deep}
Lisa Anne~Hendricks, Subhashini Venugopalan, Marcus Rohrbach, Raymond Mooney,
  Kate Saenko, Trevor Darrell, Junhua Mao, Jonathan Huang, Alexander Toshev,
  Oana Camburu, et~al.
\newblock {Deep Compositional Captioning: Describing Novel Object Categories
  without Paired Training Data}.
\newblock In {\em CVPR}, 2016.

\bibitem{antol2015vqa}
Stanislaw Antol, Aishwarya Agrawal, Jiasen Lu, Margaret Mitchell, Dhruv Batra,
  C Lawrence~Zitnick, and Devi Parikh.
\newblock {{VQA:} Visual Question Answering}.
\newblock In {\em ICCV}, 2015.

\bibitem{auer2007dbpedia}
S{\"o}ren Auer, Christian Bizer, Georgi Kobilarov, Jens Lehmann, Richard
  Cyganiak, and Zachary Ives.
\newblock {DBpedia: A Nucleus for A Web of Open Data}.
\newblock In {\em {The Semantic Web}}, pages 722--735. Springer, 2007.

\bibitem{bollacker2008freebase}
Kurt Bollacker, Colin Evans, Praveen Paritosh, Tim Sturge, and Jamie Taylor.
\newblock {Freebase: A Collaboratively Created Graph Database for Structuring
  Human Knowledge}.
\newblock In {\em ACM SIGMOD International Conference on Management of Data},
  2008.

\bibitem{chung2014empirical}
Junyoung Chung, Caglar Gulcehre, KyungHyun Cho, and Yoshua Bengio.
\newblock {Empirical Evaluation of Gated Recurrent Neural Networks on Sequence
  Modeling}.
\newblock {\em arXiv preprint arXiv:1412.3555}, 2014.

\bibitem{Imagenet}
Jia Deng, Wei Dong, Richard Socher, Li-Jia Li, Kai Li, and Li Fei-Fei.
\newblock {ImageNet: A Large-Scale Hierarchical Image Database}.
\newblock In {\em CVPR}, 2009.

\bibitem{fellbaum1998wordnet}
Christiane Fellbaum.
\newblock {\em {WordNet: An Electronic Lexical Database}}.
\newblock Bradford Books, 1998.

\bibitem{frome2013devise}
Andrea Frome, Greg~S Corrado, Jon Shlens, Samy Bengio, Jeff Dean, Tomas
  Mikolov, et~al.
\newblock {DeViSE: A Deep Visual-Semantic Embedding Model}.
\newblock In {\em NIPS}, 2013.

\bibitem{fukui2016multimodal}
Akira Fukui, Dong~Huk Park, Daylen Yang, Anna Rohrbach, Trevor Darrell, and
  Marcus Rohrbach.
\newblock {Multimodal Compact Bilinear Pooling for Visual Question Answering
  and Visual Grounding}.
\newblock In {\em EMNLP}, 2016.

\bibitem{goyal2017making}
Yash Goyal, Tejas Khot, Douglas Summers-Stay, Dhruv Batra, and Devi Parikh.
\newblock {Making the V in VQA Matter: Elevating the Role of Image
  Understanding in Visual Question Answering}.
\newblock In {\em CVPR}, 2017.

\bibitem{gupta2017aligned}
Tanmay Gupta, Kevin Shih, Saurabh Singh, and Derek Hoiem.
\newblock {Aligned Image-Word Representations Improve Inductive Transfer across
  Vision-Language Tasks}.
\newblock In {\em CVPR}, 2017.

\bibitem{he2016deep}
Kaiming He, Xiangyu Zhang, Shaoqing Ren, and Jian Sun.
\newblock {Deep Residual Learning for Image Recognition}.
\newblock In {\em CVPR}, 2016.

\bibitem{hu2018learning}
Hexiang Hu, Wei-Lun Chao, and Fei Sha.
\newblock {Learning Answer Embeddings for Visual Question Answering}.
\newblock In {\em CVPR}, 2018.

\bibitem{huang2015learning}
Sheng Huang, Mohamed Elhoseiny, Ahmed Elgammal, and Dan Yang.
\newblock {Learning Hypergraph-Regularized Attribute Predictors}.
\newblock In {\em CVPR}, 2015.

\bibitem{johnson2017clevr}
Justin Johnson, Bharath Hariharan, Laurens van~der Maaten, Li Fei-Fei,
  C~Lawrence Zitnick, and Ross Girshick.
\newblock {CLEVR: A Diagnostic Dataset for Compositional Language and
  Elementary Visual Reasoning}.
\newblock In {\em CVPR}, 2017.

\bibitem{kafle2018dvqa}
Kushal Kafle, Brian Price, Scott Cohen, and Christopher Kanan.
\newblock {DVQA: Understanding Data Visualizations via Question Answering}.
\newblock In {\em CVPR}, 2018.

\bibitem{kim2016hadamard}
Jin-Hwa Kim, Kyoung-Woon On, Woosang Lim, Jeonghee Kim, Jung-Woo Ha, and
  Byoung-Tak Zhang.
\newblock {Hadamard Product for Low-Rank Bilinear Pooling}.
\newblock In {\em ICLR}, 2016.

\bibitem{kiros2015skip}
Ryan Kiros, Yukun Zhu, Ruslan Salakhutdinov, Richard~S Zemel, Antonio Torralba,
  Raquel Urtasun, and Sanja Fidler.
\newblock {Skip-Thought Vectors}.
\newblock In {\em NIPS}, 2015.

\bibitem{krishna2017visual}
Ranjay Krishna, Yuke Zhu, Oliver Groth, Justin Johnson, Kenji Hata, Joshua
  Kravitz, Stephanie Chen, Yannis Kalantidis, Li-Jia Li, David~A Shamma, et~al.
\newblock {Visual Genome: Connecting Language and Vision using Crowdsourced
  Dense Image Annotations}.
\newblock {\em IJCV}, 123(1):32--73, 2017.

\bibitem{krizhevsky2012imagenet}
Alex Krizhevsky, Ilya Sutskever, and Geoffrey~E Hinton.
\newblock {Imagenet Classification with Deep Convolutional Neural Networks}.
\newblock In {\em NIPS}, 2012.

\bibitem{openimages}
Alina Kuznetsova, Hassan Rom, Neil Alldrin, Jasper Uijlings, Ivan Krasin, Jordi
  Pont-Tuset, Shahab Kamali, Stefan Popov, Matteo Malloci, Tom Duerig, and
  Vittorio Ferrari.
\newblock {The Open Images Dataset V4: Unified Image Classification, Object
  Detection, and Visual Relationship Detection at Scale}.
\newblock {\em arXiv preprint arXiv:1811.00982}, 2018.

\bibitem{lampert2014attribute}
Christoph~H Lampert, Hannes Nickisch, and Stefan Harmeling.
\newblock {Attribute-Based Classification for Zero-Shot Visual Object
  Categorization}.
\newblock {\em TPAMI}, 2014.

\bibitem{larochelle2008zero}
Hugo Larochelle, Dumitru Erhan, and Yoshua Bengio.
\newblock {Zero-Data Learning of New Tasks}.
\newblock In {\em AAAI}, 2008.

\bibitem{textblob}
Steven Loria.
\newblock {{TextBlob: Simplified Text Processing}}.
\newblock \url{http://textblob.readthedocs.io/en/dev/}, 2018.

\bibitem{lu2018neural}
Jiasen Lu, Jianwei Yang, Dhruv Batra, and Devi Parikh.
\newblock {Neural Baby Talk}.
\newblock In {\em CVPR}, 2018.

\bibitem{malinowski2014multi}
Mateusz Malinowski and Mario Fritz.
\newblock {A Multi-World Approach to Question Answering about Real-World Scenes
  Based on Uncertain Input}.
\newblock In {\em NIPS}, 2014.

\bibitem{noh2016image}
Hyeonwoo Noh, Paul Hongsuck~Seo, and Bohyung Han.
\newblock {Image Question Answering using Convolutional Neural Network with
  Dynamic Parameter Prediction}.
\newblock In {\em CVPR}, 2016.

\bibitem{papadopoulos2016we}
Dim~P Papadopoulos, Jasper~RR Uijlings, Frank Keller, and Vittorio Ferrari.
\newblock {We Don’t Need No Bounding-Boxes: Training Object Class Detectors
  Using Only Human Verification}.
\newblock In {\em CVPR}, 2016.

\bibitem{papadopoulos2017extreme}
Dim~P Papadopoulos, Jasper~RR Uijlings, Frank Keller, and Vittorio Ferrari.
\newblock {Extreme Clicking for Efficient Object Annotation}.
\newblock In {\em ICCV}, 2017.

\bibitem{pennington2014glove}
Jeffrey Pennington, Richard Socher, and Christopher Manning.
\newblock {Glove: Global Vectors for Word Representation}.
\newblock In {\em EMNLP}, 2014.

\bibitem{ren2015faster}
Shaoqing Ren, Kaiming He, Ross Girshick, and Jian Sun.
\newblock {Faster R-CNN: Towards Real-Time Object Detection with Region
  Proposal Networks}.
\newblock In {\em NIPS}, 2015.

\bibitem{teney2017tips}
Damien Teney, Peter Anderson, Xiaodong He, and Anton van~den Hengel.
\newblock {Tips and Tricks for Visual Question Answering: Learnings from the
  2017 Challenge}.
\newblock In {\em CVPR}, 2018.

\bibitem{teney2016zero}
Damien Teney and Anton van~den Hengel.
\newblock {Zero-Shot Visual Question Answering}.
\newblock {\em arXiv preprint arXiv:1611.05546}, 2016.

\bibitem{venugopalan2017captioning}
Subhashini Venugopalan, Lisa Anne~Hendricks, Marcus Rohrbach, Raymond Mooney,
  Trevor Darrell, and Kate Saenko.
\newblock {Captioning Images With Diverse Objects}.
\newblock In {\em CVPR}, 2017.

\bibitem{vinyals2015pointer}
Oriol Vinyals, Meire Fortunato, and Navdeep Jaitly.
\newblock {Pointer Networks}.
\newblock In {\em NIPS}, 2015.

\bibitem{wang2017fvqa}
Peng Wang, Qi Wu, Chunhua Shen, Anthony Dick, and Anton van~den Hengel.
\newblock {FVQA: Fact-Based Visual Question Answering}.
\newblock {\em TPAMI}, 2017.

\bibitem{wang2017vqa}
Peng Wang, Qi Wu, Chunhua Shen, and Anton van~den Hengel.
\newblock {The VQA-Machine: Learning How to Use Existing Vision Algorithms to
  Answer New Questions}.
\newblock In {\em CVPR}, 2017.

\bibitem{wu2016ask}
Qi Wu, Peng Wang, Chunhua Shen, Anthony Dick, and Anton van~den Hengel.
\newblock {Ask Me Anything: Free-Form Visual Question Answering based on
  Knowledge from External Sources}.
\newblock In {\em CVPR}, 2016.

\bibitem{xian2017zero}
Yongqin Xian, Bernt Schiele, and Zeynep Akata.
\newblock {Shot Learning-The Good, the Bad and the Ugly}.
\newblock In {\em CVPR}, 2017.

\bibitem{yang2016stacked}
Zichao Yang, Xiaodong He, Jianfeng Gao, Li Deng, and Alex Smola.
\newblock {Stacked Attention Networks for Image Question Answering}.
\newblock In {\em CVPR}, 2016.

\bibitem{yao2017incorporating}
Ting Yao, Yingwei Pan, Yehao Li, and Tao Mei.
\newblock {Incorporating Copying Mechanism in Image Captioning for Learning
  Novel Objects}.
\newblock In {\em CVPR}, 2017.

\bibitem{zhu2016visual7w}
Yuke Zhu, Oliver Groth, Michael Bernstein, and Li Fei-Fei.
\newblock {Visual7W: Grounded Question Answering in Images}.
\newblock In {\em CVPR}, 2016.

\end{thebibliography}
\bibliographystyle{ieee}

\end{document}